\documentclass[10pt, a4paper]{article}
\usepackage{lrec}
\usepackage{graphicx}
\usepackage{tabularx}
\usepackage{listings}
\usepackage{soul}
\usepackage{amsmath}
\usepackage{xcolor}
\usepackage{todonotes}

\usepackage{epstopdf}
\usepackage[utf8]{inputenc}

\usepackage{hyperref}
\usepackage{xstring}

\title{Give your Text Representation Models some Love: the Case for Basque
}

\name{Rodrigo Agerri$^{1*}$, Iñaki San Vicente$^{2*}$, Jon Ander Campos$^{1*}$, Ander Barrena$^{1*}$,\\
        \large{\textbf{Xabier Saralegi$^2$, Aitor Soroa$^1$, Eneko Agirre$^1$ }}}

\address{\\$^1$ IXA Group, HiTZ Centre, University of the Basque Country (UPV/EHU), $^2$Elhuyar Foundation  \\
         \{rodrigo.agerri, jonander.campos, ander.barrena, a.soroa, e.agirre\}@ehu.eus\\
         \{i.sanvicente, x.saralegi\}@elhuyar.eus\\}

\abstract{
Word embeddings and pre-trained language models allow to build rich representations of text and have enabled improvements across most NLP tasks. Unfortunately they are very expensive to train, and many small companies and research groups tend to use models that have been pre-trained and made available by third parties, rather than building their own. This is suboptimal as, for many languages, the models have been trained on smaller (or lower quality) corpora. In addition, monolingual pre-trained models for non-English languages are not always available. At best, models for those languages are included in multilingual versions, where each language shares the quota of substrings and parameters with the rest of the languages. This is particularly true for smaller languages such as Basque. In this paper we show that a number of monolingual models (FastText word embeddings, FLAIR and BERT language models) trained with larger Basque corpora produce much better results than publicly available versions in downstream NLP tasks, including topic classification, sentiment classification, PoS tagging and NER. This work sets a new state-of-the-art in those tasks for Basque. All benchmarks and models used in this work are publicly available.
\\ \newline 
}

\begin{document}

\maketitleabstract

\section{Introduction}\label{sec:introduction}

Word embeddings and pre-trained language models allow to build rich representations of text and have enabled improvements across most NLP tasks. The most successful models include word embeddings like FastText \cite{fasttext1_bojanowski2017enriching}, and, more recently, pre-trained language models which can be used to produce contextual embeddings or directly fine-tuned for each task. Good examples of the later are character-based models like Flair \cite{akbik2018coling} and masked language models like BERT \cite{devlin2019bert}. In many cases the teams that developed the algorithms also release their models, which facilitates both reproducibility and their application in downstream tasks. Given that the models are expensive to train, many companies and research groups tend to use those models, rather than building their own. This could be suboptimal as, for many languages, the models have been trained on easily obtained public corpora, which tends to be smaller and/or of lower quality that other existing corpora. In addition, pre-trained language models for non-English languages are not always available. In that case, only multilingual versions are available, where each language shares the quota of substrings and parameters with the rest of the languages, leading to a decrease in performance \cite{devlin2019bert}.  The chances for smaller languages, as for instance Basque, seem even direr, as easily available public corpora is very limited, and the quota of substrings depends on corpus size.

As an illustration of the issues mentioned above, the multilingual BERT which Basque shares with other 103 languages is based on Wikipedia corpora, where English amounts to 2.5 thousand million tokens whereas for Basque it contains around 35 million tokens. Our corpus uses, in addition to the Basque Wikipedia, corpora crawled from news outlets (191M tokens). Another important issue is subword tokenization. Thus, for some common Basque words such as \texttt{etxerantz} (to the house) or \texttt{medikuarenera} (to the doctor), the subword tokenization generated by the monolingual BERT we trained will substantially differ from the output produced by the multilingual BERT:

\begin{verbatim}
 mBERT: Et #xer #ant #z
 ours:  Etxera #ntz

 mBERT: Medi #kua #rene #ra
 ours:  Mediku #aren #era
\end{verbatim}

More specifically, mBERT's subwords tend to be shorter and less interpretable, while our subwords are closer to linguistically interpretable strings, like \texttt{mediku} (doctor) \texttt{aren} ('s) and \texttt{era} (to the).

Furthermore, most of the time the released models have been thoroughly tested only in English. Alternatively, multilingual versions have been tested in transfer learning scenarios for other languages, where they have not been compared to monolingual versions \cite{devlin2019bert}.
The goal of this paper is to compare publicly available models for Basque with analogous models which have been trained with a larger, better quality corpus. This has been possible for the FastText and Flair models. In the case of BERT, only the multilingual version for 104 languages is available for Basque. We focus on four downstream NLP tasks, namely, topic classification, sentiment classification, Part-of-Speech (POS) tagging and Named Entity Recognition (NER).

The main contribution of our work is threefold: (1) We show that, in the case of Basque, publicly available models can be outperformed by analogous models trained with appropriate corpora. (2) We improve the state of the art in four tasks for Basque. (3) We make all our models publicly available\footnote{\scriptsize \url{http://ixa2.si.ehu.es/text-representation-models/basque}}. Our results show that this is indeed the case for FastText and Flair models. In the case of BERT, we show that the monolingual Basque BERTeus model\footnote{Also available from Hugginface in \scriptsize{\url{https://huggingface.co/ixa-ehu/berteus-base-cased}}} outperforms the official multilingual BERT by a large margin, with up to 10 absolute point improvement. The increases are consistent for the three models in all four tasks. Our experiments have been performed training models which are analogous to the corresponding existing pre-trained models. This allows to make a head-to-head comparison, and to conclude that the training corpus is key. In the case of BERT, training a monolingual model instead of a multilingual one shared with 103 other languages is also a key factor, so we can rule out that it is not any change in the models that cause the better performance.

Although our results set a new state of the art for the four tasks, our goal was not to train and use the systems in the most optimal manner. Instead, and as mentioned before, we have focused on head-to-head comparisons. In this sense, better results could be expected using variations of pre-trained models that have reported better results for English \cite{liu2019roberta}, or making an effort in developing better adaptations to each of the tasks.


\section{Related work}\label{sec:related-work}


Deep learning methods in NLP rely on the ability to represent words as continuous vectors on a low dimensional space, called word embeddings. Word2vec~\cite{mikolov2013distributed} or GloVe~\cite{Pennington14glove:global} are among the best models that build word embeddings by analyzing co-occurrence patterns extracted from large corpora. FastText~\cite{fasttext1_bojanowski2017enriching} proposes an improvement over those models, consisting on embedding subword units, thereby attempting to introduce morphological information. Rich morphology languages such as Basque should especially profit from such word representations. FastText distributes embeddings for more than 150 languages trained on Common Crawl and  Wikipedia. In this paper we build FastText embeddings using a carefully collected corpus in Basque and show that it performs better than the officially distributed embeddings in all NLP we tested, which stresses the importance of a following a carefully designed method when building and collecting the corpus.

The aforementioned methods generate static word embeddings, that is, they provide a unique vector-based representation for a given word independently of the context in which the word occurs. Thus, if we consider the Basque word \emph{banku}\footnote{In English: bank.}, static word embedding approaches will calculate one vector irrespective of the fact that the same word \emph{banku} may convey different senses when used in different contexts, namely, ``financial institution'',``bench'', ``supply or stock'', among others. In order to address this problem, contextual word embeddings are proposed; the idea is to be able to generate different word representations according to the context in which the word appears. Examples of such contextual representations are ELMO \cite{Peters:2018} and Flair \cite{akbik2018coling}, which are built upon LSTM-based architectures and trained as language models. More recently, \newcite{devlin2019bert} introduced BERT, a model based on the transformer architecture trained as a masked language model, which has obtained very good results on a variety of NLP tasks. The multilingual counterpart of BERT, called mBERT, is a single language model pre-trained from corpora in more than 100 languages. mBERT enables to perform transfer knowledge techniques among languages, so that systems can be trained on datasets in languages different to the one used to fine tune them~\cite{heinzerling-strube-2019-sequence,DBLP:conf/acl/PiresSG19}.

When pre-training mBERT the corpora sizes in different languages are very diverse, with English corpora being order of magnitudes larger than that of the minority languages. The authors alleviate this issue by oversampling examples of lower resource languages. However, as the parameters and vocabulary is shared across the languages, this means that each language has to share the quota of substrings and parameters with the rest of the languages. As shown in the introduction, this causes tokenization problems for low resource languages.

CamemBERT~\cite{martin2019camembert} is a recent effort to build custom RoBERTa \cite{liu2019roberta} pre-trained language models for French, using 138GB of text from a monolingual French corpus extracted from Common Crawl. The authors show that CamemBERT obtains significant improvements on many French tasks compared to the publicly available multilingual BERT. However, the architecture used in CamemBERT is different to BERT, which makes the comparison among the models less straightforward. In contrast, we try to mimic the BERT architecture and parameters when building our BERTeus models, thus allowing a head-to-head comparison among them. In addition we also report results for FastText and Flair models.

\section{Building Basque models}\label{sec:build-basq-models}


In order to train accurately word embeddings and language models we need a corpus as large as possible, but also meeting certain criteria. Those criteria included clean and correctly structured text (having paragraph and document limits, specially relevant for training BERT language models). With that aim, we collected a corpus in Basque comprising crawled news articles from online newspapers and the Basque Wikipedia\footnote{The dump from 2019/10/01 was used.}. In total the collected corpus contains 224.6 million tokens, of which 35 million come from the Wikipedia. Table \ref{tab:corpus} shows the composition of the corpus. Henceforth, we will refer to this corpus as the \emph{Basque Media Corpus} (BMC).

\begin{table}[!ht]\small
\centering
\begin{tabular}{@{\hspace{0.3cm}}lcc} \hline
 {\textbf{}} & {\textbf{Text type}} & {\textbf{Million tokens}}  \\ \hline
Wikipedia & enciclopedia & 35M \\
Berria newspaper & news & 81M  \\
EiTB & news & 28M \\
Argia magazine & news & 16M \\
Local news sites & news &  64.6M \\ \hline
\hline
\textbf{BMC} & & 224.6M\\
\hline
\end{tabular}
\caption{Composition of the Basque Media Corpus (BMC).}\label{tab:corpus}
\end{table}


\subsection{Static Embeddings: FastText}\label{sec:build-basq-models:static}

To the best of our knowledge, the only publicly available static word embeddings for Basque are those distributed by Facebook and learned using an approach based on the skipgram model, where  each  word  is  represented as a bag of character n-grams \cite{fasttext1_bojanowski2017enriching}. The approach is implemented in the FastText library\footnote{\scriptsize \url{https://fasttext.cc/}}, and two types of pre-trained word vectors are available (based on 300 dimensions):

\textbf{Wiki word vectors} (FastText-official-wikipedia) were trained on Wikipedia using the skip-gram model described in \cite{fasttext1_bojanowski2017enriching} with default parameters (a window size of 5, 3-6 length character n-grams and 5 negatives).

\textbf{Common Crawl word vectors} (FastText-official-common-crawl) were trained on Common Crawl and Wikipedia using CBOW with position-weights, with character n-grams of length 5, a window size 5 and 10 negatives\cite{fasttext2_grave2018learning}.

In this work, we trained \textbf{BMC word vectors} (FastText-BMC) of 300 dimensions on the BMC corpus described above, using CBOW with position-weights and the default parameters of the original paper. \cite{fasttext1_bojanowski2017enriching}.

\subsection{Contextual String Embeddings: Flair}\label{sec:build-basq-models:flair}

Flair refers to both a deep learning system and to a specific type of character-based contextual word embeddings. Flair (embeddings and system) have been successfully applied to sequence labeling tasks obtaining state-of-the-art results for a number of English Named Entity Recognition (NER) and Part-of-Speech tagging benchmarks \cite{akbik2018coling}, outperforming other well-known approaches such as BERT and ELMO \cite{devlin2019bert,Peters:2018}. In any case, Flair is of interest to us because they distribute their own Basque pre-trained embedding models obtained from a corpus of 36M tokens (combining OPUS and Wikipedia).

\paragraph{Flair-BMC models:} We train our own Flair embeddings using the BMC corpus with the following parameters: Hidden size 2048, sequence length of 250, and a mini-batch size of 100. The rest of the parameters are left in their default setting. Training was done for 5 epochs over the full training corpus. The training of each model took 48h on a Nvidia Titan V GPU.

\paragraph{Flair Embeddings:} Flair's embeddings model words as sequences of characters. Moreover, the vector-based representation of a word will depend on its surrounding context. More specifically, to generate word embeddings they feed sentences as sequences of characters into a character-level Long short-term memory (LSTM) model which at each point in the sequence is trained to predict the next character. Given a sentence, a forward LSTM language model processes the sequence from the beginning of the sentence to the last character of the word we are modeling extracting its output hidden state. A backward LSTM performs the same operation going from the end of the sentence up to the first character of the word. In this case, the extracted hidden state contains information propagated from the end of the sentence to this point. Both hidden states are concatenated to generate the final embedding.

\paragraph{Pooled Contextualized Embeddings:} Flair embeddings, however, struggle to generate an appropriate word representation for words in underspecified contexts, namely, in sentences in which, for example, local information is not sufficient to know the named entity type of a given word. In order to address this issue a variant of the original Flair embeddings is proposed: ``Pooled Contextualized Embeddings'' \cite{akbik2019naacl}. In this approach, every contextualized embedding is kept into a \emph{memory} which is later used in a pooling operation to obtain a global word representation consisting of the concatenation of all the local contextualized embeddings obtained for a given word. They reported significant improvements for NER by using this new version of Flair embeddings. Note that this does not affect to the way the Flair pre-trained embedding models are calculated. The pooling operation is involved in the process of using such pre-trained models in order to obtain word representations for a given task such as NER or POS using the Flair system.

\paragraph{Flair System:} For sequence labeling tasks, the calculated character-based embeddings are passed into a BiLSTM-CRF system based on the architecture proposed by \cite{huang2015bidirectional}. For text classification tasks, the computed Flair embeddings are fed into a BILSTM\footnote{A single layer of size 128 was used, with word reprojection and a dropout value of 0.3068.} to produce a document level embedding which is then used in a linear layer to make the class prediction. Although for best results they recommend to stack their own Flair embeddings with additional static embeddings such as FastText, in this paper our objective is to compare the official pre-trained Flair embeddings for Basque with our own Flair-BMC embeddings.

\subsection{BERT language models}\label{sec:build-basq-models:bert}

We have trained a BERT \cite{devlin2019bert} model for Basque Language using the BMC corpus motivated by the rather low representation this language has in the original multilingual BERT model. In this section we describe the methods used for creating the vocabulary, the model architecture, the pre-training objective and procedure.

The main differences between our model and the original implementation are the corpus used for the pre-training, the algorithm for sub-word vocabulary creation and the usage of a different masking strategy that is not available for the BERT\textsubscript{BASE} model yet.

\paragraph{Sub-word vocabulary}

We create a cased sub-word vocabulary containing 50,000 tokens using the unigram language model based sub-word segmentation algorithm proposed by \newcite{kudo2018subword}. We do not use the same algorithm as BERT because the WordPiece \cite{wu2016google} implementation they originally used is not publicly available. We have increased the vocabulary size from 30,000 sub-word units up to 50,000 expecting to be beneficial for the Basque language due to its agglutinative nature. Our vocabulary is learned from the whole training corpus but we do not cover all the characters in order to avoid very rare ones. We set the coverage percentage to $99.95$.

\paragraph{Model Architecture}

In the same way as the original BERT architecture proposed by \newcite{devlin2019bert} our model is composed by stacked layers of Transformer encoders \cite{vaswani2017attention}. Our approach follows the BERT\textsubscript{BASE} configuration containing 12 Transformer encoder layers, a hidden size of 768 and 12 self-attention heads for a total of 110M parameters.

\paragraph{Pre-training objective}

Following BERT original implementation, we train our model on the Masked Language Model (MLM) and Next Sentence Prediction (NSP) tasks. Even if the necessity of the NSP task has been questioned by some recent works \cite{yang2019xlnet,liu2019roberta,lample2019cross} we have decided to keep it as in the original paper to allow for head-to-head comparison. For the MLM, given an input sequence composed of N tokens $x_1, x_2, ..., x_n$ we select a $15\%$ of them as masking candidates. Then, $80\%$ of these selected tokens are masked by replacing them with the [MASK] token, $10\%$ are replaced with a random word of the vocabulary and the remaining $10\%$ are left unchanged. In order to create input examples for the NSP task, we take two segments, $A$ and $B$, from the training corpus, where $B$ is the true next segment for $A$ only for $50\%$ of the cases. For the rest, $B$ is just a random segment from the corpus. At the end, the model is trained to optimize the sum of the means of the MLM and NSP likelihoods.

As our vocabulary consists of sub-word units, we use whole-word masking (WWM), that applies the masking to whole words instead of sub-word units. This new masking strategy makes the MLM task more difficult for the system as it has to predict the whole word instead of predicting just part of it. An upgraded version of BERT\textsubscript{LARGE}\footnote{\scriptsize{\url{https://github.com/google-research/bert}}} has proven that WWM has substantial benefits in comparison with previous masking that was done after the sub-word tokenization.

\paragraph{Pre-training procedure}

Similar to \cite{devlin2019bert} we use Adam with learning rate of $1e-4$, $\beta_1=0.9$, $\beta_2=0.999$, L2 weight decay of $0.01$, learning rate warmup over the first $10,0000$ steps, and linear decay of the learning rate. The dropout probability is fixed to $0.1$ on all the layers.

As the attentions are quadratic to the sequence length, making longer sequences much more expensive, we pre-train the model with sequence length of $128$ for $90\%$ of the steps and sequence length of $512$ for $10\%$ of the steps. In total we train for $1,000,000$ steps and a batch size of $256$. The first $90\%$ steps are trained using Cloud v2 TPUs and for the rest of the steps we use Cloud v3 TPUs \footnote{\scriptsize{\url{https://cloud.google.com/tpu/pricing}}}.


\section{Evaluation and Results}\label{sec:eval-results}








We conduct an extensive evaluation on four well known NLP tasks: Topic Classification, Sentiment Classification, Part-of-Speech (POS) tagging and Named Entity Recognition (NER). The datasets used for each task are described in their respective sections.

We train our systems to perform the following comparisons: (i) FastText official models (Wikipedia and Common Crawl) vs FastText-BMC model; (ii) the official Flair embedding models vs our Flair-BMC model and, (iii) BERTeus with respect to multilingual BERT.

To train the Flair system we use the parameters specified in \cite{akbik2019naacl} for Pooled Contextual Embeddings. Flair is tuned on the development data using the test only for the final evaluation. We do not use the development set for training.

For comparison between BERT models we fine-tune on the training data provided for each of the four tasks with both the official multilingual BERT \cite{devlin2019bert} model and with our BERTeus model (trained as described in Section \ref{sec:build-basq-models:bert}).

Every reported result for every system is the average of five randomly initialized runs. The POS and NER experiments using mBERT and BERTeus are performed using the transformers library \cite{Wolf2019HuggingFacesTS} where it is recommended to remove the seed for random initialization.

\subsection{Topic Classification}\label{sec:topic}

For the task of topic classification a dataset containing 12k news headlines (brief article descriptions) was compiled from the Basque weekly newspaper Argia\footnote{\scriptsize \url{https://www.argia.eus}}. News are classified uniquely according to twelve thematic categories. The dataset was divided into train (70\%), development (15\%) and test datasets (15\%). The Basque Headlines Topic Classification (BHTC) dataset is publicly available under CC license.\footnote{\scriptsize \url{https://hizkuntzateknologiak.elhuyar.eus/assets/files/BHTC.tgz}}

The FastText and Flair related experiments were conducted using the Flair text classifier with the same hyperparameter settings\footnote{max-epoch 50, learning rate 0.1, minibatch size 64, and  patience 3 }. BERTeus and mBERT fine-tuning was performed over 3 epochs with a learning rate of 2e-5 and a batch size of 16.

\begin{table}[!ht]\small
\centering
\begin{tabular}{@{\hspace{0.3cm}}lcc} \hline
 {\textbf{}} & {\textbf{Micro F1}} &  {\textbf{Macro F1}} \\ \hline
\textbf{Static Embeddings} & & \\
FastText-Wikipedia & 65.00 & 54.98 \\
FastText-Common-Crawl & 28.82 & 3.73  \\
FastText-BMC  & 69.45 & 60.14 \\
\hline
\textbf{Flair Embeddings}\\
Flair-official & 65.25 & 55.83 \\
Flair-BMC  & 68.61 & 59.38  \\ \hline
\textbf{BERT Language Models}\\
mBERT-official  & 68.42 & 48.38  \\
BERTeus  & \textbf{76.77}	& \textbf{63.46}  \\
\hline
\textbf{Baseline} \\
TF-IDF Logistic Regression & 63.00 & 49.00 \\
\hline
\end{tabular}
\caption{Basque topic classification results on Argia corpus.}\label{tab:topic}
\end{table}

Table \ref{tab:topic} shows the results obtained by the different models. Firstly, it can be seen that every BMC-trained model outperforms its official counterpart for all the three settings, by at least 4\% in all cases, for both micro and macro F1-scores. The best results are obtained by BERTeus. For each type of embedding or language models, the results show the effectiveness of developing your own models for your own language.

\subsection{Sentiment Classification}\label{sec:polarity}

Sentiment classification is evaluated using a corpus of tweets containing messages related to the cultural domain\footnote{\scriptsize{\url{https://hizkuntzateknologiak.elhuyar.eus}}}. The corpus contains annotations for three classes (positive, negative and neutral), and a total of 2,936 examples. For the experiments in this paper the corpus was divided into train (80\%), test (10\%) and development (10\%) sets. Class distribution corresponds to 32\% positive, 54\% neutral and 14\% negative.

For FastText and Flair experiments we used the Flair text classifier with the same hyperparameter settings as those used for Topic Detection in Section \ref{sec:topic}. This is also the case for BERTeus and mBERT: fine-tuning over 3 epochs with a learning rate of 2e-5 and a batch size of 16.

\begin{table}[!ht]\small
\centering
\begin{tabular}{@{\hspace{0.3cm}}lcc} \hline
 {\textbf{}} & {\textbf{micro F1}} &  {\textbf{Macro F1}} \\ \hline
\textbf{Static Embeddings} & & \\
FastText-Wikipedia & 71.10 &	66.72 \\
FastText-Common-Crawl & 66.16 & 58.89  \\
FastText-BMC  & 72.19 &	68.14 \\
\hline
\textbf{FlairEmbeddings}\\
Flair-official & 72.74 & 67.73 \\
Flair-BMC  & 72.95	& 69.05 \\ \hline
\textbf{BERT Language Models}\\
mBERT-official  & 71.02 & 66.02 \\
BERTeus  & \textbf{78.10}	& \textbf{76.14} \\
\hline
\textbf{Baseline} \\
SVM \cite{san2019multilingual} & 74.02 & 69.87\\ \hline
\end{tabular}
\caption{Basque sentiment classification task on Behagune tweet corpus.}\label{tab:sentiment}
\end{table}

Table \ref{tab:sentiment} shows that every BMC-trained model outperforms its official counterpart for all the three settings. The difference between Flair-official and Flair-BMC is rather small in terms of micro F1-score, but Flair-BMC obtains a higher macro F1. Furthermore, BERTeus establishes a new state-of-the-art on this particular dataset, improving previous results \cite{san2019multilingual} by 4.08 points in micro F1 score, and 6.27 in macro F1 score. BERTeus specially improves the classification of the positive and negative classes, which provides a boost in terms of macro F1.

\subsection{POS Tagging}\label{sec:pos-tagging}

In order to facilitate comparison with previous state-of-the-art methods, we experiment with the Universal Dependencies 1.2 data, which provides train, development and test partitions. The Basque UD treebank \cite{aranzabe2015automatic} is based on a conversion from part of the Basque Dependency Treebank (BDT) \cite{aduriz2003construction}. The treebank consists of 5274 sentences (60563 tokens) and covers mainly literary and journalistic texts. Previous best result for this dataset in Basque has been reported by \cite{heinzerling-strube-2019-sequence}. They use Byte Pair Embeddings (BPEmb) which are trained on subword-segmented text. The segmentation of the text is obtained by using the Byte Pair Encoding (BPE) unsupervised algorithm, which iteratively merges frequent pairs of adjacent symbols into new ones. The BPEmbeddings are then fed into a LSTM sequence tagging architecture.


\begin{table}[!ht]\footnotesize
\centering
\begin{tabular}{@{\hspace{0.3cm}}lccc} \hline
 \textbf{} & \textbf{Word Accuracy} \\ \hline
\textbf{Static Embeddings} & \\
FastText-Wikipedia & 94.09 \\
FastText-Common-Crawl & 91.95 \\
FastText-BMC  & 96.14 \\ \hline
\textbf{FlairEmbeddings} \\
Flair-official & 97.50 \\
Flair-BMC  &  97.58 \\ \hline
\textbf{BERT Language Models}\\
mBERT-official &  96.37 \\
BERTeus & \textbf{97.76} \\ \hline
\textbf{Baseline} \\
\cite{heinzerling-strube-2019-sequence} & 96.10 \\ \hline
\end{tabular}
\caption{Basque POS tagging results on UD 1.2.}\label{tab:pos}
\end{table}

Table \ref{tab:pos} shows the results obtained by the different models. Firstly, it can be seen that every BMC-trained model outperforms its official counterpart for all the three settings, albeit the difference between Flair-official and Flair-BMC is rather small. Secondly, our BMC-models (both BERTeus and Flair) establish a new state-of-the-art on this particular dataset significantly improving over the result reported by \cite{heinzerling-strube-2019-sequence}. In any case, the results show the effectiveness of developing your own models for your own language. This is especially supported by the difference in performance obtained by BERTeus with respect to mBERT-official.

\begin{table}[!ht]\footnotesize
\centering
\begin{tabular}{@{\hspace{0.3cm}}lccc} \hline
 \textbf{} &\textbf{Precision} & \textbf{Recall} & \textbf{F1} \\ \hline
\textbf{Static Embeddings} & & &  \\
FastText-Wikipedia & 72.42 & 50.28 & 59.23 \\
FastText-Common-Crawl & 72.09 & 45.31 & 55.53 \\
FastText-BMC  & 74.12 & 67.33 & 70.56 \\
\hline
\textbf{Flair embeddings}\\
Flair-official & 81.86 & 79.89 & 80.82 \\
Flair-BMC & 84.32 & 82.66 & 83.48 \\ \hline
\textbf{BERT Language Models} \\
mBERT-official  & 81.24 & 81.80 & 81.52 \\
BERTeus  & 87.95 & 86.11 & \textbf{87.06} \\ \hline
\textbf{Baseline} \\
\cite{agerri2016robust} & 80.66 & 73.14 & 76.72 \\ \hline
\end{tabular}
\caption{Basque NER results on EIEC corpus.}\label{tab:ner}
\end{table}

\begin{table*}[!t]
\centering
\begin{tabular}{@{\hspace{0.3cm}}lcccc} \hline
\textbf{} & \multicolumn{4}{c}{\textbf{Task}} \\ 
 & {\textbf{Topic Classification}} & {\textbf{Sentiment}} &  {\textbf{POS}} & {\textbf{NER}}\\ \hline
\multicolumn{5}{@{}l}{\textbf{Static Embeddings}} \\
FastText-Wikipedia & 65.00 & 71.10 & 94.09 & 59.23 \\
FastText-Common-Crawl & 28.82 & 66.16 & 91.95 & 55.53 \\
FastText-BMC  & 69.45 & 72.19 & 96.14 & 70.56 \\
\hline
\multicolumn{5}{@{}l}{\textbf{Flair Embeddings}}\\
Flair-official & 65.25 & 72.74 & 97.50 & 80.82 \\
Flair-BMC  & 68.61 & 72.95 & 97.58 & 83.48 \\ \hline
\multicolumn{5}{@{}l}{\textbf{BERT Language Models}} \\
mBERT-official  & 68.42 & 71.02 & 96.37 & 81.52 \\
BERTeus  & \textbf{76.77} & \textbf{78.10} & \textbf{97.76} & \textbf{87.06} \\ \hline
\textbf{Baselines} & 63.00 & 74.02 & 96.10 & 76.72 \\ \hline
\end{tabular}
\caption{Summary table across all tasks. Micro F1 scores are reported.}\label{sec:results-discussion:table}
\end{table*}

\subsection{Named Entity Recognition}\label{sec:named-entity-recogn}

EIEC\footnote{\scriptsize{\url{http://ixa2.si.ehu.eus/eiec/eiec_v1.0.tgz}}} is a gold standard corpus for NER in Basque \cite{alegria2006lessons}. The corpus contains 44K tokens for training (3817 unique entities) and 15K for testing (931 entities). Although EIEC is annotated with four entity types (Location, Person, Organization and Miscellaneous), the \emph{Miscellaneous} class is rather sparse, occurring only in a proportion of 1 to 10 with respect to the other three classes. Thus, in the training data there are 156 entities annotated as \emph{Miscellaneous} whereas for each of the other three classes it contains around 1200 entities.

As for the previous tasks above, we compare our BMC-trained models (FastText, Flair and BERT) with respect to the official releases. We also compare with the previous published baseline on this dataset, which trains a Perceptron model with a simple and shallow feature set combined with clustering features based on unigram matching \cite{agerri2016robust}.

Table \ref{tab:ner} reports the results of our comparison for NER using the usual conlleval script from the CoNLL NER shared tasks \footnote{\scriptsize{\url{https://www.clips.uantwerpen.be/conll2002/ner/}}}. We can see that the BMC-trained models improve over the official models distributed by FastText, Flair and mBERT, with a much larger margin than for POS tagging. It is also worth mentioning that there is a nice balance between the precision and the recall obtained by both Flair-BMC and BERTeus. Finally, we think that the larger differences in performance between NER and POS tagging might be partially due to the small size of the EIEC NER corpus. In any case, more experimentation is required to clarify the issue.


\section{Discussion}\label{sec:discussion}

As illustrated by the overview of results provided by Table \ref{sec:results-discussion:table}, our BMC-trained models improve over previous work for all four tasks evaluated. This is true when we compare each model with their official counterpart, but also in absolute terms, where BERTeus outperforms every other model by a wide margin.

We believe that the POS tagging results on UD 1.2 are especially interesting, given that previous work had already established a very competitive baseline \cite{heinzerling-strube-2019-sequence} and that our aim was to provide a head-to-head comparison between analogous models. For example, best performance with Flair is usually the result of combining Flair string contextual embeddings with static embeddings such as FastText \cite{akbik2018coling}, but our objective was not to optimize the experiments for best results but, rather, to compare with the Flair-official model for Basque. In the same way, when training BERTeus we strived to comply with the BERT architecture and parameters thus allowing a
head-to-head comparison between them.

Another interesting issue that arises by looking at the scores across tasks is the robust performance of the FastText-BMC embeddings. In fact, the FastText-BMC embeddings used to train the Flair document classifier outperform multilingual BERT for the Topic Detection and Sentiment Analysis tasks. Moreover, and unlike the FastText-official models, the results of FastText-BMC do not fluctuate as much, performing quite robustly across tasks. Finally, our Flair-BMC model outperforms both Flair-official and multilingual BERT in every setting, although it remains far from the performance displayed by BERTeus.

All these issues show the importance of carefully collecting a corpus to generate pre-trained language models specific to each language, especially for less-resourced ones which are usually under-represented in large multilingual models.

In the case of BERT, we see an additional factor which, for sure, influences the better results of the in-house BERTeus model, that of training a monolingual model instead of a highly multilingual one. In this sense, we think that the good performance of BERTeus is due to the combination of two factors: larger and better corpus, as well as the fact of using language-specific subword tokenization. These improvements provide a huge boost in performance which, for example, help to correctly classify difficult NER examples (such as those of metonymy), as displayed in Table \ref{tab:ner-examples}. In this example, ``United States'' and ``Western Europe'' are correctly classified as ORGANIZATION entities whereas mBERT predicts them to be LOCATION, which is incorrect in this particular context.

\begin{table*}[!t]
\centering
\resizebox{\textwidth}{!}{
\begin{tabular}{lllllllllllllll}
      & Chechenya &  &   &  &  &  & United & States &  & Western & Europe &  &  &  \\
      & Txetxeniako & gai & horretan & agerian & geratu & da & Estatu & Batuek & eta & Mendebaldeko & Europak & eman & dioten & sustengua. \\
\textbf{Gold: }      & B-LOC & O & O & O & O & O & \textbf{B-ORG} & \textbf{I-ORG} & O & \textbf{B-ORG} & \textbf{I-ORG }& O & O & O. \\
\textbf{mBERT} & B-LOC & O & O & O & O & O & \color{red} B-LOC & \color{red} I-LOC & O & \color{red} B-LOC & \color{red} I-LOC & O & O & O. \\
\textbf{BERTeus} & B-LOC & O & O & O & O & O & \textbf{B-ORG} & \textbf{I-ORG} & O & \textbf{B-ORG} & \textbf{I-ORG} & O & O & O. \\
\end{tabular}}
\caption{An example from the EIEC corpus where the mBERT model misses to disambiguate the cases of metonymy. English translation of the sentence: \textit{``On that issue about Chechenya the support given by the United States and Western Europe has been visible.''}}
\label{tab:ner-examples}
\end{table*}

\section{Conclusions and future work}\label{sec:concl-future-work}

In this paper we show that a number of Basque monolingual models trained on our own corpora produce much better results than publicly available versions in downstream NLP tasks. More specifically, we build monolingual word embeddings using FastText, and pre-trained language models using Flair and BERT. All of them have been trained with the Basque Media Corpus (BMC), a 226M token corpus comprising Wikipedia and news outlets. The results on a broad range of NLP tasks, including topic classification, sentiment classification, PoS tagging and NER, show improvements in all tasks with respect to third-party models. The best results in all cases are obtained with our BERTeus model, with up to 10 absolute point increases with regard to the use of the official multilingual BERT, setting a new state-of-the-art in all four tasks.

It is important to note that our experiments have been performed training models which are analogous to the corresponding existing pre-trained models, instead of optimizing for best results in each of the tasks. This strategy allows to make a head-to-head comparison, and to conclude that the training corpus is key. In the case of BERT, training a monolingual model instead of a multilingual one shared with 103 other languages is also a key factor, so we can rule out that it is not changes in the models that cause the increased performance. In this sense, new variations of those models have reported better results \cite{liu2019roberta}. In the future, we plan to release models which try to keep up with the state-of-the-art results of the time.

Our results for Basque seem to indicate that, specially in the case of small languages, it is crucial to devote the necessary resources to gather large high-quality corpora and to train monolingual models on those corpora.

In order to facilitate reproducibility all benchmarks and models used in this work are publicly available. We also hope that these models will serve to improve the results of NLP applications for Basque, both at research centres and industry.


\section{Acknowledgments}

This work has been partially funded by the Basque Government (IXA excellence research group and Elesight, Hazitek grant no. ZL-2018/00787), the~Spanish Ministry of Science, Innovation and Universities (DeepReading RTI2018-096846-B-C21, MCIU/AEI/FEDER, UE) and by \textit{Ayudas Fundación BBVA a Equipos de Investigación Científica 2018} (BigKnowledge).  Jon Ander Campos enjoys a doctoral grant from the Spanish MECD. Ander Barrena enjoys a post-doctoral grant ESPDOC18/101 from the UPV/EHU. Rodrigo Agerri is funded by the Ramon y Cajal Fellowship RYC-2017-23647 and acknowledges the~support of the NVIDIA Corporation with the~donation of a Titan V GPU used for this research. We also acknowledge the support of Google Cloud.

\vspace{0.5cm}

\noindent \textbf{Author Contributions:} The first four authors (marked with $^{*}$) contributed equally to this work.

\section{Bibliographical References}

\bibliographystyle{lrec}
\bibliography{main}

\end{document}